\documentclass[11pt,a4paper]{article}
\usepackage[hyperref]{acl2020}
\usepackage{times}
\usepackage{latexsym}
\usepackage[all]{xy}

\usepackage{tikz-cd}

\usepackage[british]{babel}
\usepackage{mathtools,amsfonts,amssymb,colortbl,etoolbox,booktabs}

\usepackage[hang,flushmargin]{footmisc}

\newcommand{\RR}{\mathbb{R}}
\newcommand{\topspace}[1]{\mathcal #1}
\newcommand{\TPS}[2]{\mathrm{TPS}_{#1}(#2)}
\newcommand{\nbd}[2]{\mathcal{N}_{#1}{(#2)}} 
\newcommand{\nnbd}[2]{\mathcal{N}'_{#1}{(#2)}}

\newcolumntype{L}{>{$}l<{$}} 
\newcolumntype{R}{>{$}r<{$}} 
\newcolumntype{C}{>{$}c<{$}} 

\usepackage{microtype}

\aclfinalcopy 
\def\aclpaperid{151} 

\title{Topology of Word Embeddings:  Singularities Reflect Polysemy}

\author{Alexander Jakubowski \\
  Heinrich Heine University\\
  Düsseldorf\\
  \texttt{jakubowskialexander}\\
  \texttt{@gmail.com} \\\And
  Milica Gašić\\
  Heinrich Heine University\\
  Düsseldorf\\
  \texttt{gasic}\\
  \texttt{@hhu.de} \\\And
  Marcus Zibrowius\\
  Heinrich Heine University\\
  Düsseldorf\\
  \texttt{marcus.zibrowius}\\
  \texttt{@cantab.net} \\
  }
\date{}

\begin{document}
\maketitle
\begin{abstract}
The manifold hypothesis suggests that word vectors live on a submanifold within their ambient vector space.  We argue that we should, more accurately, expect them to live on a \emph{pinched} manifold:  a singular quotient of a manifold obtained by identifying some of its points. The identified, singular points correspond to polysemous words, i.e.\ words with multiple meanings.  Our point of view suggests that monosemous and polysemous words can be distinguished based on the topology of their neighbourhoods.  We present two kinds of empirical evidence to support this point of view:  
(1) We introduce a topological measure of polysemy based on persistent homology that correlates well with the actual number of meanings of a word.
(2) We propose a simple, topologically motivated solution to the SemEval-2010 task on \textit{Word Sense Induction \& Disambiguation} that produces competitive results.
\end{abstract}

\section{Introduction}
\label{sec:intro}
\blfootnote{%
This work is licensed under a Creative Commons Attribution 4.0 International Licence. Licence details: \url{http://creativecommons.org/licenses/by/4.0/}.
}

Static word embeddings attempt to represent words by vectors in a high-dimensional vector space \(\RR^n\) in such a way that words of similar meaning are represented by (cosine) similar vectors, and vice versa.
According to the manifold hypothesis, we should expect these vectors to lie within a lower-dimensional \textbf{word space} \(\topspace W\), a subspace of \(\RR^n\) that resembles a manifold.  To what extent this hypothesis is true in this and other contexts is the subject of ongoing research \cite{fefferman:testing-manifold-hypothesis}.  In this paper, we argue and demonstrate that for the word space \(\topspace W\), polysemy is a principal obstruction to any strict interpretation of the manifold hypothesis.

That polysemy presents a serious obstacle to the creation of adequate word vector representations is clear from the outset.  Take, for example, a polysemous word like ``mole''.  We would want the vectors representing ``birthmark'' and ``counterspy'' to be similar to the vector of ``mole'', but \emph{not} similar to each other.  This is impossible.  In order for similarity of vectors to accurately encode similarity in meaning, we need vectors representing meanings, not words.


Let us therefore hypothesize a \textbf{space of meanings} \(\topspace M\) that accurately represents all possible meanings and their similarities.  Our argument is a simple topological observation based on the relationship between this space \(\topspace M\) and the word space \(\topspace W\).  For an idealised language, where there is a bijection between meanings and words, these two spaces would agree.  For a natural language, however, multiple points of \(\topspace M\) get identified with a single point of \(\topspace W\).  This process corresponds to a topological construction that we refer to as \textbf{pinching} (see Figure~\ref{fig:pinched-space}). 
It is easy to see that a space resulting from pinching cannot be a manifold.  Thus, even if the space of meanings \(\topspace M\) satisfies the manifold hypothesis perfectly, the pinched space \(\topspace W\) cannot satisfy the hypothesis near polysemous words.\footnote{
    It may appear that a similar complication arises from synonyms, multiple words with a single meaning.  However, synonyms are irrelevant for our analysis; see the discussion at the end of Section~\ref{sec:word-space-as-pinched-manifold}.
}

Based on this intuition, and using tools from Topological Data Analysis, we introduce a  measure for the polysemy of a word based on its vector embedding.  Our experiments show that this \textbf{topological polysemy} (TPS) correlates well with the actual number of meanings that a word has.  In addition, we present an approach to the SemEval-2010 task on \emph{Word Sense Induction \& Disambiguation} (task 14) \cite{semeval2010-14}.   This approach is independent of TPS, but based on the same ideas. Despite its simplicity, it is almost on par with the best performing algorithm within the 2010 challenge, and outperforms far more complicated approaches.  

We see these experimental results as strong evidence that our interpretation of the word space \(\topspace W\) as a pinched manifold is more adequate than a more naïve view of \(\topspace W\) as an actual manifold.


\section{Background}
\subsection{Topology}
\label{sec:topology}
\begin{figure}
\centering
\newcommand{\mfdlabel}[1]{(#1)}
\begin{tabular}{ccccc}
\raisebox{-.45\height}{\includegraphics[width=0.22\linewidth]{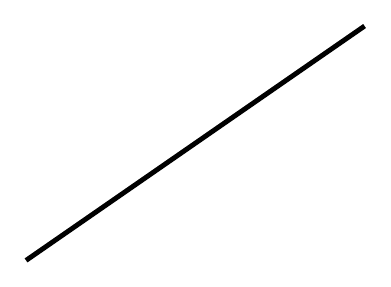}} & $\cong$ &
\raisebox{-.45\height}{\includegraphics[width=0.22\linewidth]{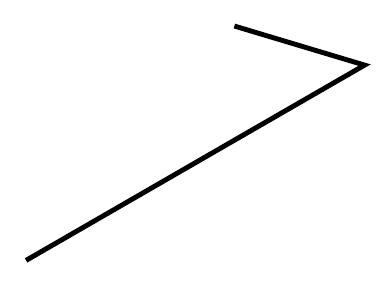}} & $\cong$ &
\raisebox{-.45\height}{\includegraphics[width=0.22\linewidth]{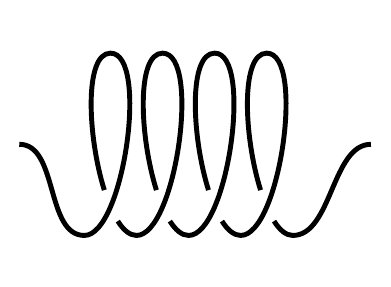}}
\\
\mfdlabel{a} && \mfdlabel{b} && \mfdlabel{c}
\\[4ex]
\raisebox{-.45\height}{\includegraphics[width=0.2\linewidth]{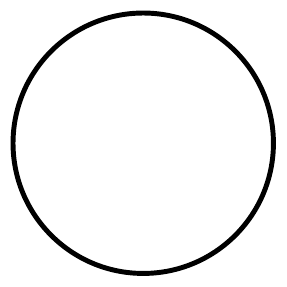}} & $\cong$ &
\raisebox{-.45\height}{\includegraphics[width=0.2\linewidth]{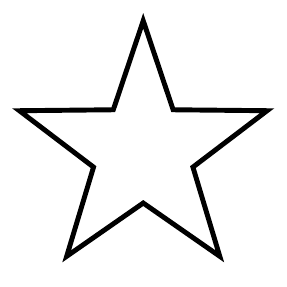}} & $\cong$ &
\raisebox{-.45\height}{\includegraphics[width=0.2\linewidth]{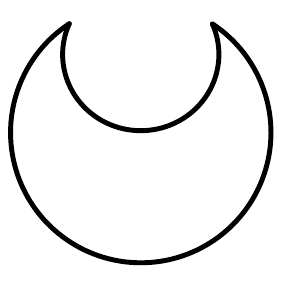}}
\\
\mfdlabel{d} && \mfdlabel{e} && \mfdlabel{f}
\end{tabular}
\vspace{4ex}

\begin{tabular}{ccc}
\raisebox{-.45\height}{\includegraphics[width=0.4\linewidth]{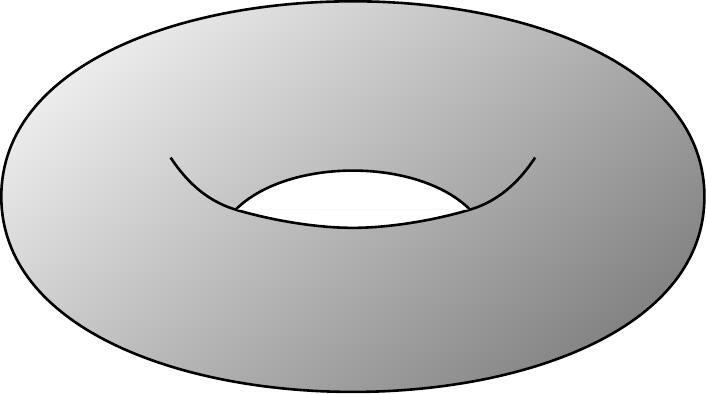}} & $\cong$ &
\raisebox{-.45\height}{\includegraphics[width=0.4\linewidth]{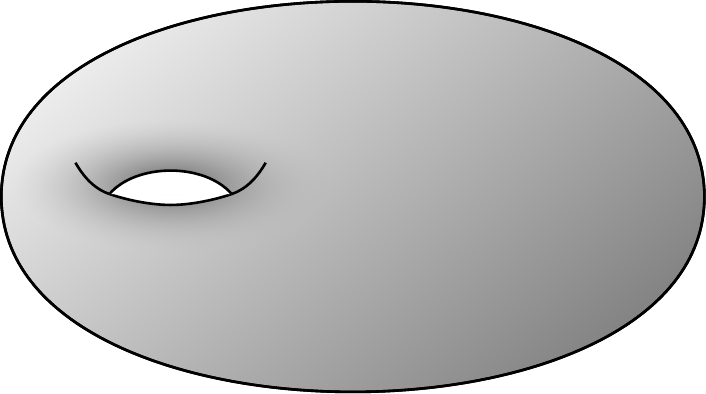}} 
\\[5ex]
\mfdlabel{g} && \mfdlabel{h}
\\[4ex]
\raisebox{-.45\height}{\includegraphics[width=0.2\linewidth]{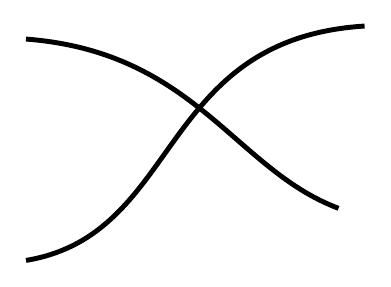}} & &
\raisebox{-.45\height}{\includegraphics[width=0.4\linewidth]{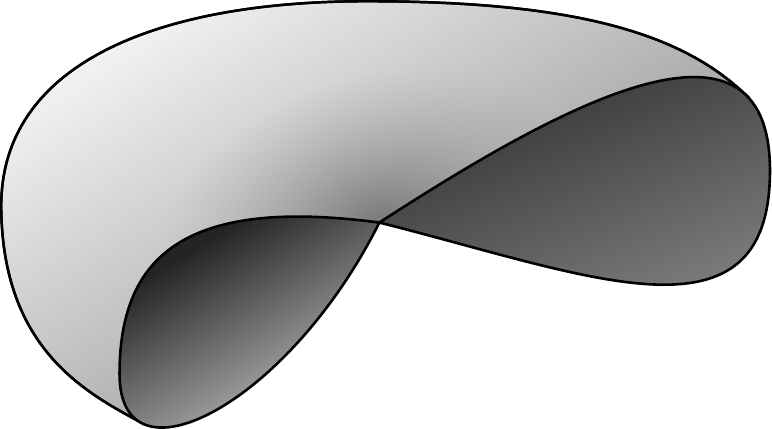}}
\\
\mfdlabel{i} && \mfdlabel{j}
\end{tabular}
    \caption{Some subspaces of \(\RR^3\): various deformations of an open line segment (a, b, c), deformations of a circle (d, e, f), a torus (g) and a deformation of the torus (h), two intersecting line segments (i), and a surface with a figure eight as boundary (j)}
    \label{fig:spaces}
\end{figure}

A space, for us, is a topological space.  
Readers unfamiliar with the notion may simply think of metric spaces, or indeed of subspaces of euclidean space \(\RR^n\).   
Two such spaces are considered equivalent, or \textbf{homeomorphic}, if they can be deformed into each other.  We will not make this precise here, but we hope that Figure~\ref{fig:spaces}, in which homeomorphic spaces are connected by the symbol ``\(\cong\)'', gives a clear intuition.
As flexible as this notion may appear, the deformations considered do keep certain properties of a space invariant.  Crucially, homeomorphic spaces always have the same number of connected components and the same number of holes.  Topologists have developed a myriad of more subtle invariants that allow us to decide whether two spaces are homeomorphic.  We refer to \citet{hatcher:AT} for an introduction into this vast field.

Two kinds of spaces will be important for us:  manifolds and pinched manifolds.  A (topological) \textbf{manifold} is a space in which each point has a neighbourhood homeomorphic to an open ball of \(\RR^d\) for some \(d\) \citep[cf.][\S\,3.3]{hatcher:AT}.\footnote{
    Manifolds are moreover required to be \textit{Hausdorff}, a technical condition that all metric spaces satisfy.
}
We call \(d\) the local dimension of the manifold at that point. The spaces (a), (b) and (c) in Figure~\ref{fig:spaces} are manifolds since each point has a neighbourhood homeomorphic to an open interval in \(\RR^1\), and so are the spaces (d), (e), (f).  The spaces (g) and (h) are manifolds because each point has a neighbourhood homeomorphic to an open disk in \(\RR^2\).   Space (i), on the other hand, is not a manifold, because the point of intersection has no neighbourhood homeomorphic to an open ball of any dimension, and neither is space (j), because the manifold condition is violated at all boundary points.  The spaces ``without corners'', i.e.\ examples (a), (c), (d), (g) and (h) in Figure~\ref{fig:spaces} are not only topological manifolds but even \emph{differentiable} manifolds, but this distinction will be of no importance to us.

\begin{figure}
    \centering
    \raisebox{-.45\height}{\includegraphics[width=0.4\linewidth]{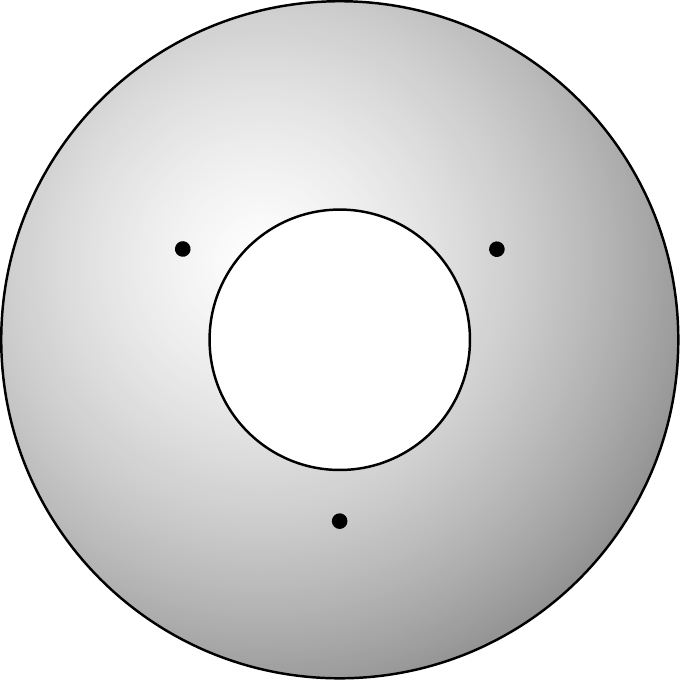}}
    $\,\leadsto\,$
    \raisebox{-.45\height}{\includegraphics[width=0.4\linewidth]{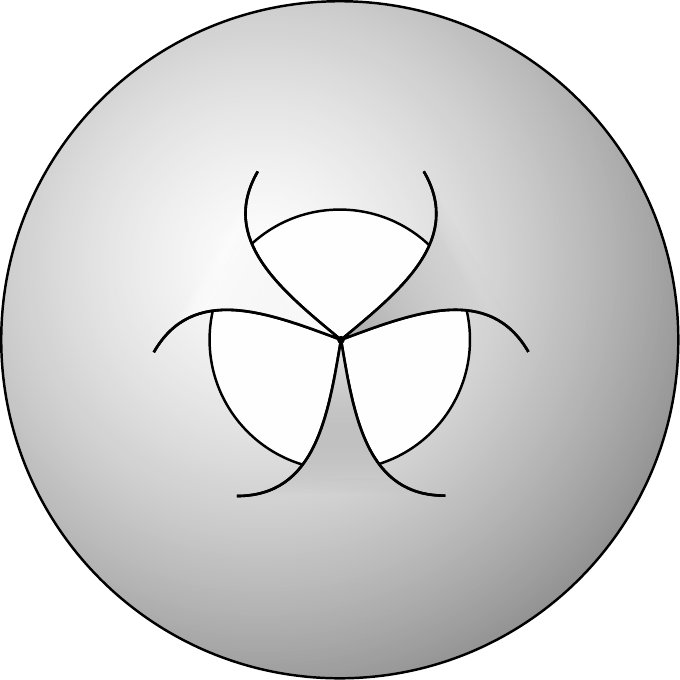}}
    \caption{The effect of pinching on the torus (example (g) from Figure~\ref{fig:spaces}):  before \textit{(left)} and after the identification of three marked points to a singular point \textit{(right)}}
    \label{fig:pinched-space}
\end{figure}

By a \textbf{pinched manifold}, we will mean a space obtained from a manifold by marking a finite number of points in different colours, and identifying (``glueing together'') all points of the same colour, as illustrated in Figure~\ref{fig:pinched-space}.  In a pinched manifold, the neighbourhoods of most points still look like open balls, but the neighbourhoods of the identified points look like several balls glued together at their centres.  We will call these identified points \textbf{singular points}.   

Singular points can thus easily be distinguished from non-singular points by the topology of their neighbourhoods.  
More precisely, we can distinguish them by counting the number of connected components of their \textbf{punctured neighbourhoods}: neighbourhoods of a point from which the point itself has been removed.  The punctured neighbourhood of a non-singular point is a single punctured ball, and thus is connected, at least in dimensions \(d\geq 2\). The punctured neighbourhood of a singular point obtained by identifying \(k>1\) points, on the other hand, is a disjoint union of \(k\) punctured balls, and thus has several connected components.  Thus, in dimensions \(d\geq 2\), a point is singular if and only if small punctured neighbourhoods of it have more than one connected component.  

Puncturing the neighbourhood, i.e.\ removing the centre, is crucial for this distinction.  The unpunctured neighbourhoods of singular and non-singular points are not distinguishable by the usual topological invariants.  (In technical terms, the neighbourhoods of both types of points are \textit{contractible}.)

\subsection{Topological data analysis}
\label{sec:TDA}
\begin{figure}
    \centering
    \includegraphics[width=\linewidth]{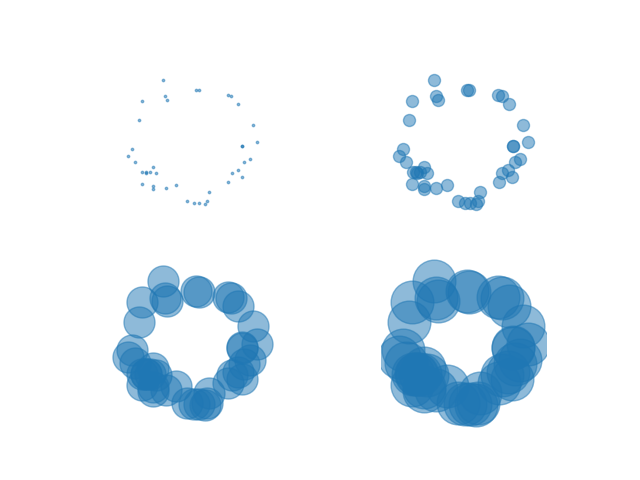}
    \caption{Points noisily sampled from the unit circle \textit{(top left)} and the corresponding spaces \(\topspace W_r\) for different radii \(r\)}
    \label{fig:tda-cloud}
\end{figure}

Topological data analysis (TDA) is an instrument for extracting topological information from a point cloud, that is a finite set of vectors $\topspace W_0 = \{ p_1, p_2, \dots, p_N\} \subset \mathbb{R}^n$.
The point cloud itself is trivial from a topological point of view. The fundamental assumption of TDA is that the vectors of $\topspace W_0$ are not randomly distributed but instead are sampled from some underlying space $\topspace W\subset \RR^n$ which -- unlike the point cloud itself -- is topologically interesting.  A human immediately recognises that the points in Figure~\ref{fig:tda-cloud} have been sampled from a circle.  TDA provides algorithms that encode this intuition, and extend it to higher dimensions.

One such algorithm is \textbf{persistent homology}. The basic idea is to replace $\topspace W_0$ with the union $\topspace W_r$ of all open balls of a certain radius $r$ centred at the points of $\topspace W_0$. As we vary this radius, we obtain a sequence of spaces, 
starting for $r=0$ with the point cloud itself and ending at some high value of $r$ with a space in which all balls are merged into a single big blob.  We compute certain topological invariants, the so-called Betti numbers $b_i$, for each space $\topspace W_r$.  The Betti number \(b_i\) counts certain \(i\)-dimensional features of the space. For example, $b_0$ is the number of connected components and $b_1$ is the ``number of holes''; both are equal to $1$ for the two spaces in the lower half of Figure~\ref{fig:tda-cloud}.   

\begin{figure}
    \centering
    \includegraphics[width=\linewidth]{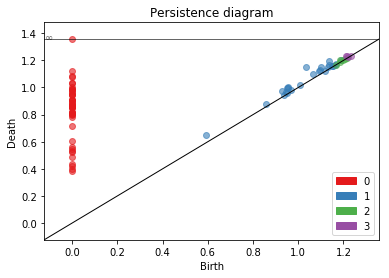}
    \caption{An example of a persistence diagram, summarizing the persistent homology of some point cloud \(\topspace W_0\) in degrees \(i=0\), \(1\), \(2\) and \(3\).   Each dot encodes the life span of a distinct feature.  Features of different degrees are displayed in different colours, as indicated in the lower right corner.  For the computations in this paper, we will focus on the degree zero features, i.e.\ on connected components, indicated in red.  As all of these are already present in the point cloud \(\topspace W_0\), they all have horizontal coordinate equal to zero.  Their vertical coordinates are the radii at which different components merge 
    }
    \label{fig:pers-diag}
\end{figure}

The radii at which different \(i\)-dimensional features appear and disappear can be summarized into a multiset and visualized as a two-dimensional \textbf{persistence diagram} \(D\) as in Figure~\ref{fig:pers-diag}.  Each dot in this diagram encodes the life span of a distinct feature: its horizontal coordinate is the \emph{smallest} radius \(r\) at which the feature is present in \(\topspace W_r\), its vertical coordinate is the \emph{largest} radius \(r\) at which it is present. 
Points that lie far off the diagonal correspond to features that \textit{persist} across a wide range of values of $r$, and are hence likely to reflect features of the underlying space $\topspace W$. 
For technical reasons, 
every point on the diagonal is also included in the persistence diagram \(D\) with infinite multiplicity.

The \textbf{Wasserstein distance} provides a notion of distance between two such persistence diagrams, and hence a measure of similarity between different point clouds and their underlying spaces. For two diagrams $D$ and $\tilde{D}$ it is defined as:
\begin{equation*}
  W ( D, \tilde{D} ) := \underset{\eta: D \rightarrow \tilde{D}}{\inf}\bigg( \underset{x \in D}{\sum} \| x - \eta (x)\|_{\infty}\bigg) 
\end{equation*}
where $\eta$ runs over all bijections between the two diagrams. As all points on the diagonal are included in both diagrams, such bijections always exist.

The computation of persistent homology can be restricted to a range of degrees $i$.  In this paper, we will concentrate on persistent homology in degree $i=0$, which is essentially a systematic application of single-linkage clustering.  Computations in higher dimensions quickly become very expensive. For an in-depth discussion of the concepts mentioned in this section we recommend \cite{CompTop:Intro}.

\subsection{Word vector embeddings}
\label{sec:word-vectors}
The distributional hypothesis states that ``the meaning of words lies in their use''~\cite{witt53}. This provides the basis for distributional semantics, a data driven study of word meanings.  Words are modelled as vectors in such a way that (cosine) similarity of vectors corresponds to similarity in the distributions of the corresponding words in natural language, and hence to semantic similarity.  In the most naïve approaches, the dimension of these vectors corresponds to the number of distinct words in the language.  More sophisticated implementations in which word vectors are real-valued but of significantly smaller dimension are popularly known as \textbf{word vector embeddings}.
They have proven important for various tasks of natural language processing~\cite{collobert2011natural,Lubis_Heck_van_Niekerk_Gasic_2020}. 

Early word vector embeddings were constructed in latent semantic analysis using singular value decomposition.  Neural methods were introduced by \citet{bengio2003neural}, and popularised by the algorithms word2vec~\cite{mikolov2013efficient,mscc13} and GloVe~\cite{pennington2014glove}.  
Our method of choice in this paper is fastText~\cite{bojanowski2017enriching}, which can produce high-quality embeddings from relatively small corpora.  All of these methods produce \textbf{static} embeddings:  they assign to each word a single, context-independent vector.

There is, of course, a lot of existing and ongoing research to overcome the difficulties inherent in adequately representing polysemous words.
One way to address polysemy is to produce multiple, context-dependent embeddings for the same word. The deep learning approaches mentioned above are amenable to this by incorporating heuristics~\cite{wordrep-ACL12} or non-parametric clustering~\cite{neelakantan-etal-2014-efficient}.  More recently, transformer based models that exploit massive datasets have been used to produce contextualised word embeddings.  Examples of these are CoVe~\cite{mccann2017learned}, ELMo~\cite{Peters:2018}, and BERT~\cite{devlin-etal-2019-bert} and its variants ERNIE \cite{sun2019ernie} and RoBERTa \cite{roberta}. 
Alternative approaches address polysemy by training multi-lingual word embeddings on multi-lingual corpora~\cite{dufter-etal-2018-embedding,heyman-etal-2019-learning}. 

As the problem of polysemy is, at least partially, resolved in all of these more advanced approaches, we would expect the phenomenon studied in this paper to be less pronounced in the embeddings they produce.  We therefore concentrate exclusively on mono-lingual static embeddings.  Our analysis will not require any data beyond such an embedding.

\section{The topology of the word space}
\subsection{The word space as a pinched manifold}
\label{sec:word-space-as-pinched-manifold}
In order to explain the apparent efficiency of machine learning, the \textbf{manifold hypothesis} postulates that, in general, real world data tends to live on a small-dimensional submanifold of the vector space in which it is represented \cite{bengio:representation-learning,fefferman:testing-manifold-hypothesis}.  For word vector embeddings, the ambient space \(\RR^n\) typically has dimension \(n\) in the range \(50\leq n \leq  300\). The hypothesis states that word vectors in fact lie on, or are densely distributed around, a submanifold \(\topspace W\subset \RR^n\) of much smaller dimension.  What this hypothetical \textbf{word space} \(\topspace W\) might look like is an intriguing question.  Work of \citet{arora:linear} suggests a dimension of \(\topspace W\) as low as five.  It is easy to imagine even smaller subspaces of \(\topspace W\), like a line segment connecting ``cold'', ``cool'', ``lukewarm'', ``warm'' and ``hot'', or a circle connecting ``north'', ``east'', ``south'', ``west''. But the global structure seems mysterious.  

\begin{figure}
    \centering
    \includegraphics[width=\linewidth]{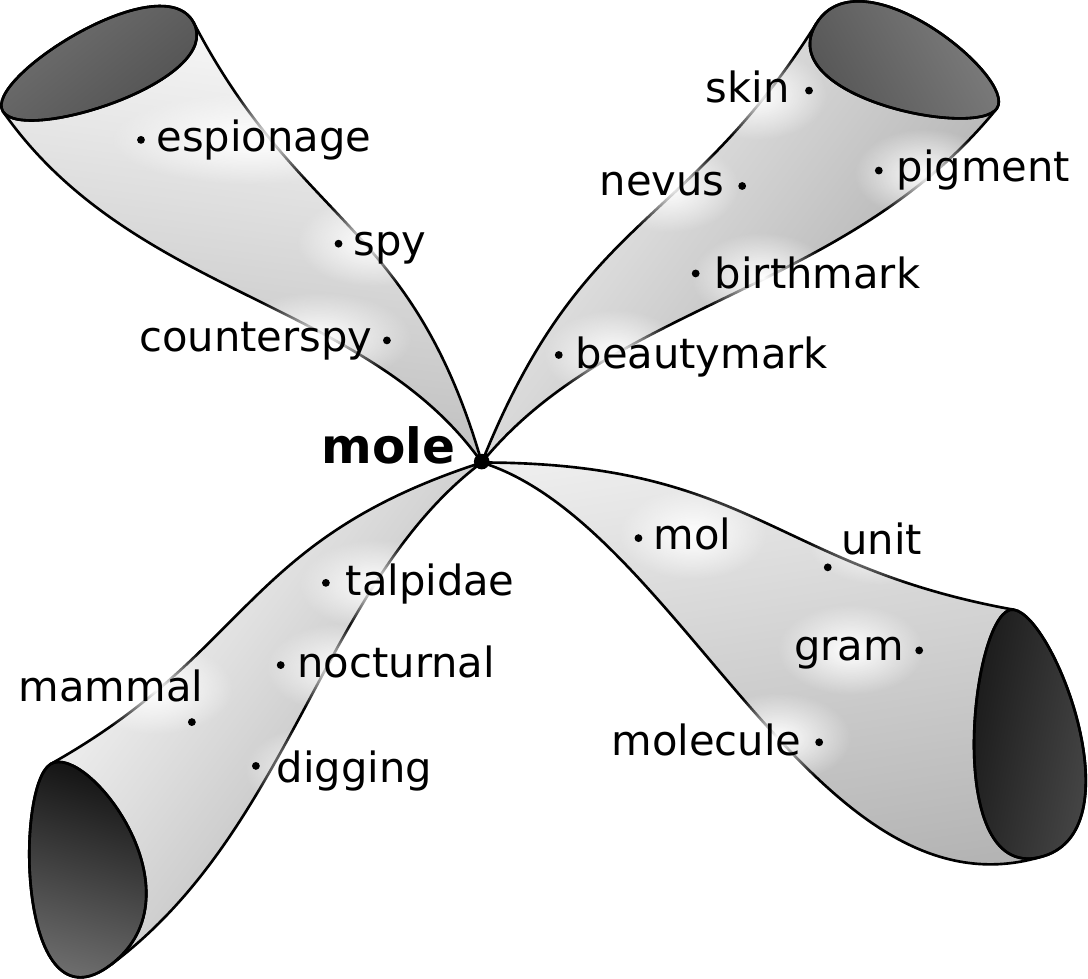}
    \caption{An idealized picture of the word space \(\topspace W\) near ``mole'': four regions of the meaning manifold are glued together to a single word}
    \label{fig:main-picture}
\end{figure}

The manifold hypothesis has two parts: (1) that \(\topspace W\) is of small dimension, and (2) that \(\topspace W\) is a manifold.\footnote{
    It may not be evident what the correct notion of ``dimension'' should be for arbitrary subspaces.  However, there are much larger classes of spaces than manifolds to which the notion of dimension extends in a straight-forward manner.
}  
It is the second statement that we would like to challenge.  We argue that, in the vicinity of polysemous words, \(\topspace W\) cannot possibly have the structure of a manifold, i.e.\ it cannot resemble an open ball of any dimension.  The best we can expect is that this might be true for some \textbf{space of meanings} \(\topspace M\) -- a space that parametrizes all possible meanings that words of a given language may assume -- from which \(\topspace W\) is obtained by identifying multiple meanings to a single word.  This identification process is precisely the pinching construction discussed in Section~\ref{sec:topology}.  For example, we should expect the neighbourhood of the polysemous word ``mole'' in \(\topspace W\) to be obtained from the neighbourhoods of its different meanings in \(\topspace M\), all glued together as in Figure~\ref{fig:main-picture}.  Thus, even if we optimistically hypothesize the space of meanings \(\topspace M\) to be a manifold, the word space \(\topspace W\) cannot be: it is at best a \emph{pinched} manifold.  It is this hypothesis that we will pursue in the following.  (If \(\topspace M\) has more complicated local structure, then \emph{a fortiori} so does \(\topspace W\).)

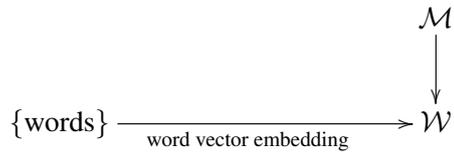
\begin{figure}
\[
    \xymatrix@C=10em{
    & \topspace M \ar[d]\\
    \{\text{words}\} \ar[r]_{\text{word vector embedding}} &\topspace W
    }
\]
\caption{The relation of the space of meanings \(\topspace M\), the space of word vectors \(\topspace W\) and the set of words of a language.  Synonyms may get identified under a given word vector embedding, symbolized here by the horizontal map.  Multiple meanings get identified to a single word vector under the vertical map}
\label{fig:synonyms}
\end{figure}

\begin{@empty}
\renewcommand{\vec}[1]{\mathbf{#1}}
The presence of synonyms in a language has no bearing on this analysis.  To explain this, we need to temporarily distinguish carefully between a word \(w\) and its associated word vector \(\vec v_w\).  The word space \(\topspace W\) should more precisely be called \emph{space of word vectors}, since this is the space in which the vectors \(\vec v_w\) live, not the words themselves. Under a given word vector embedding, synonyms \(w\) and \(w'\) may get mapped to the same word vector \(\vec v_w = \vec v_{w'}\).  However, this does not affect the relation of the space of meanings \(\topspace M\) to the space of word vectors \(\topspace W\) in any way.  The situation is summarized in Figure~\ref{fig:synonyms}.\footnote{
    It is of course debatable whether the equation \(\vec v_w = \vec v_{w'}\) would really hold for any pair of synonyms in practice.  It seems more likely that the vectors \(\vec v_w\) and \(\vec v_{w'}\) would simply lie very close together.
}
\end{@empty}

With this discussion out of the way, we will from now on again simplify our terminology by identifying words with their associated vectors, and refer to \(\topspace W\) as word space.  

\subsection{A topological measure of polysemy}
As explained at the end of Section~\ref{sec:topology}, we can distinguish a singular point of a pinched manifold from a non-singular point by counting the connected components of a small punctured neighbourhood of the point.  What is more, the number of these components reflects the number of points that were glued together in the pinching process.  Thus, according to our view of the word space \(\topspace W\) as a pinched quotient of the manifold of meanings \(\topspace M\), the number of different meanings of a word should be reflected by the number of components of a punctured neighbourhood of the word.  

Of course, the relevant number of components is not directly visible from the discrete point cloud formed by the word vectors.  Rather, the components can only be estimated by some form of clustering.  In this section, we describe a measure of the number of components based on degree zero persistent homology, as introduced in Section~\ref{sec:TDA}.  

Fix a word vector embedding, a target word \(w\), and a neighbourhood size \(n\).  As already indicated, we will abuse language by identifying a word with its vector under the embedding in the following.   The \textbf{topological polysemy} \(\TPS{n}{w}\) of \(w\) with respect to our fixed word vector embedding and our chosen neighbourhood size \(n\) is the Wasserstein norm of a normalized punctured neighbourhood of \(w\).  That is, \(\TPS{n}{w}\) is computed as follows:
\begin{enumerate}
    \item  
    Normalize all word vectors \(v\) to have $L_2$-norm \(\lVert v \rVert = 1\).
    \item
    Consider the punctured neighbourhood \(\nbd{n}{w}\) consisting of the $n$ closest neighbours of $w$, excluding $w$ itself. 
    \item
    Pass to the normalized punctured neighbourhood \(\nnbd{n}{w}\) by translating \(w\) to lie at the origin and projecting all vectors to the unit sphere:
    \[
    \nnbd{n}{w} := \left\{  \frac{v-w}{\lVert v-w \rVert} \quad\middle\vert\quad v \in \nbd{n}{w} \right\}
    \]
    \item
    Compute the degree zero persistence diagram of \(\nnbd{n}{w}\).
    \item
    \(\TPS{n}{w}\) is the Wasserstein norm of this persistence diagram, i.e.\ the Wasserstein distance between the computed and the empty persistence diagram.
\end{enumerate}
The general normalization in Step~1 is included because word embeddings are trained only on cosine similarity; the length of each vector has no apparent meaning.  The normalization allows us to compute directly with difference vectors between word vectors of high cosine similarity.  The normalization in Step~3 is included because we have fixed the \emph{cardinality} \(n\) of the neighbourhood, not its diameter.  Without any normalization in this step, we would be measuring mostly the density of the word cloud around \(w\).  The normalization by projection onto the unit sphere may seem somewhat radical, but it is topologically motivated:  the topological invariants we use cannot distinguish a punctured ball from its boundary sphere. (In technical terms, the punctured ball and its boundary are \emph{homotopy equivalent}; cf.\ \citet{hatcher:AT}, Chapter~0.)

%

\section{Empirical evidence}
We present two pieces of empirical evidence that support our view of the word space as a pinched manifold.  The experiments in Sections~\ref{sec:GS} and \ref{sec:wordnet} show that the topological polysemy defined above correlates with the actual number of meanings of a word.  In Section~\ref{sec:semeval}, we describe a simple approach to the SemEval-2010 task on word sense induction based on our topological intuition.  

\subsection{Experimental setup}
\label{sec:experimental}
All experiments are based on data provided with the SemEval-2010 task on \textit{Word Sense Induction \& Disambiguation} \cite{semeval2010-14}. The task is as follows:  
Assign a total of $8\,915$ \textbf{instances}, extracted from various sources including CNN and ABC, of $100$ different polysemous \textbf{target words} ($50$ nouns and $50$ verbs) to clusters based on their \textbf{context}, such that instances with different meanings get mapped to different clusters and instances with the same meaning get mapped to the same cluster. A context is simply a paragraph of text that the target word appears in. Figure~\ref{example:context} shows an exemplary context for an instance of the target word ``cultivate''.  Note that labels are only provided for a test set; this is an unsupervised learning task.

The training set provided comprises $65$\,M occurrences of $127\,151$ different words.  We use this corpus to train our own vector representations using the python module for fastText \cite{bojanowski2017enriching}. For the computation of the persistence diagrams and the Wasserstein distance we use the GUDHI library \cite{gudhi:urm}.  

\begin{figure}
\begin{center}
\makebox{\begin{minipage}{0.9\linewidth}
\itshape
``Don't forget the Tatun Mountains, which shelter the town. In the old days , Tanshui folk who cultivated farms on the slopes had to walk for an hour to get to their crops. These days you can take a local mini-bus.''
\end{minipage}}
\end{center}
\caption{An exemplary context of an instance of the target word ``cultivate''}
\label{example:context}
\end{figure}


\subsection{Correlation of TPS with the SemEval gold standard}
\label{sec:GS}
\begin{figure}
    \centering
    \includegraphics[width=\linewidth]{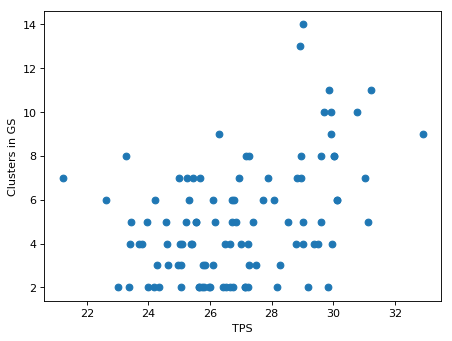}
    \caption{The topological polysemy \(\TPS{50}{w}\) plotted against the number of clusters in the SemEval gold standard, for the \(100\) SemEval target words \(w\)}
    \label{fig:corWithGS}
\end{figure}

The SemEval data set includes a gold standard 
for the \(100\) target words.  The number of clusters in this gold standard is equal to the number of true meanings of each word, as perceived by humans. Figure~\ref{fig:corWithGS} shows our measure of polysemy \(\TPS{50}{w}\) for the \(100\) target words \(w\) plotted against these cluster counts.  

Correlation coefficients between the gold standard and \(\TPS{n}{w}\) for varying neighbourhood sizes \(n\) are displayed in the first column of Table~\ref{tbl:correlations}.  We found the highest correlation for \(n=50\), equal to \(0.424\).    Neighbourhoods consisting of just ten or less words are clearly too small to capture multiple meanings.  On the other hand, for high values of \(n\), the neighbourhoods become too large to adequately reflect the local structure of the word space around the target word.  This is likewise unsurprising: recall from Section~\ref{sec:topology} that the manifold condition is a \emph{local} condition around each point.  Larger neighbourhoods of a point on a manifold can be arbitrarily complicated, and so can larger neighbourhoods of singular points on a pinched manifold.

The third column of Table~\ref{tbl:correlations} shows that \(\TPS{n}{w}\) does \emph{not} correlate with the frequency of the words in the SemEval corpus.  This is important, as frequency itself correlates with polysemy.  The  absence of correlation between \(\TPS{n}{w}\) and frequency strengthens our assertion that \(\TPS{n}{w}\) indeed measures polysemy.

\begin{table}[t]
\newcommand{\insign}[1]{\color{gray}{#1}}
\centering
\begin{tabular}{cccc}
\toprule
$n$   & {\textbf{TPS${}_n$ vs.}}       & {\textbf{TPS${}_n$ vs.}}    & {\textbf{TPS${}_n$ vs.}}      \\
& {\textbf{GS}} & {\textbf{synsets}} & {\textbf{frequency}}\\
\midrule
$10$  & \insign{\llap{--\,}0.001} & 0.122 & 0.002  \\
$40$  & 0.411  & 0.096 & \llap{--\,}0.003 \\
$50$  & 0.424  & 0.085 & \llap{--\,}0.006 \\
$60$  & 0.414  & 0.076 & \llap{--\,}0.008 \\
$100$ & 0.333  & 0.055 & \llap{--\,}0.013 \\
\midrule
\parbox{3em}{\centering \small \textit{sample size}} & $100$ & $62\,049$ & $127\,151$ \\
\bottomrule
\end{tabular}%
\caption{Correlations between \(\TPS{n}{w}\) and the number of meanings of \(w\) according to the Sem\-Eval gold standard (Section~\ref{sec:GS}), the number of WordNet synsets (Section~\ref{sec:wordnet}), and the frequency of \(w\) in the SemEval training corpus. The last line indicates the number of words on which the correlation is computed. The gray entry is not statistically significant, but all other entries are (\(p\)-value \(< 10^{-3}\))}
\label{tbl:correlations}
\end{table}
\begin{figure}[t]
    \centering
    \includegraphics[width=\linewidth]{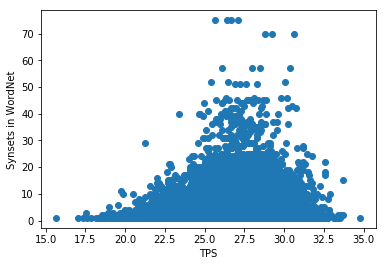}
    \caption{The topological polysemy \(\TPS{50}{w}\) plotted against the number of synsets in WordNet, for all \(62\,049\) words \(w\) in the SemEval corpus that have a WordNet entry}
    \label{fig:corWithWordNet}
\end{figure}
\begin{table}[t]
\centering
\begin{tabular}{lccc}
\toprule
       & {\textbf{GS}}     & {\textbf{WordNet}} & {\textbf{TPS$_{50}$}} \\
\midrule
sniff  & 3      & 3       & 27.262     \\
reap   & 2      & 2       & 26.658     \\
bow    & 5      & 14      & 28.533     \\
chip   & 13     & 14      & 28.910     \\
house  & 14     & 14      & 28.999     \\
\bottomrule
\end{tabular}%
\caption{Some examples of words and their corresponding cluster count in the SemEval gold standard and WordNet as well as their TPS-measure for $n=50$}
\label{tbl:GSexamples}
\end{table}

\subsection{Correlation of TPS with WordNet synsets}
\label{sec:wordnet}
The correlation with the gold standard is a good indication of the validity of our method, but it is based on just \(100\) samples.  The number of meanings, as perceived by humans, of a much larger set of words can be extracted from WordNet \cite{WordNet:Hardcopy}, specifically the number of synsets associated with each word.  Of course, we cannot expect the correlation between our topological polysemy and these numbers of synsets to be as high as for the SemEval gold standard.  Firstly, we have trained our fastText vectors specifically on the SemEval training set, which does not capture the breadth of WordNet, and which does not comprise enough data to yield adequate embeddings for non-target words.  Secondly, WordNet captures distinctions in meaning far more granular that one could hope to detect within the, say, \(50\) closest neighbours of a word.

Nonetheless, plotting \(\TPS{50}{w}\) against the number of synsets for all \(62\,049\) words of the SemEval corpus that have a WordNet entry indicates a clear trend, see Figure~\ref{fig:corWithWordNet}.  Correlation coefficients for varying \(n\) are included in Table~\ref{tbl:correlations}.

\subsection{The SemEval task}
\label{sec:semeval}
Our hypothesis that the word space is a manifold pinched at polysemous words also suggests the following, direct approach to the SemEval-2010 task itself, which we call \textbf{Overlap with Punctured Neighbourhood (OPN)}.
Fix a neighbourhood size \(n\).  In a first step, we cluster punctured neighbourhoods of size \(n\) of the \(100\) target words using a common clustering algorithm like \(k\)-means or dbscan \cite{dbscan}. The different clusters of the neighbourhood cloud obtained in this way are taken to represent different meanings of the target word.  In a second step, we assign a given instance of the target word to the cluster of the neighbourhood cloud that has the highest relative word overlap with the context of that instance.

\begin{@empty}
For clustering with dbscan, we found that the best results are achieved with parameter values \(\mathrm{Eps}=0.09\) and \(\mathrm{MinPts} = 2\)  and large neighbourhood sizes~\(n\).
\newcommand{\TPSmin}{\mathrm{TPS}_{\text{min}}}%
\newcommand{\TPSmax}{\mathrm{TPS}_{\text{max}}}%
\newcommand{\percentile}{\%}%
The \(k\)-means clustering algorithm requires the number \(k\) of clusters aimed for as a parameter. We experimented both with fixed values of \(k\) and with a word-dependent variable value \(k(w)\), predicted using TPS as follows.  Define the TPS-percentile \(\percentile(w)\) of a target word \(w\) as 
\[
    \percentile(w) := \left\lceil\frac{\TPS{50}{w}-\TPSmin}{\TPSmax-\TPSmin}\cdot 100\right\rceil,
\]
where \(\TPSmin\) and \(\TPSmax\) denote the minimum and maximum values that \(\TPS{50}{\cdot}\) assumes on all target words, respectively, and where \(\lceil\cdot\rceil\) denotes rounding to the next largest integer.  Thus, the percentile is an integer between \(0\) and \(100\) that reflects how large \(\TPS{50}{w}\) is in comparison to all other target words.  The expected number of clusters \(k(w)\) is defined as 
\[
 k(w) := \begin{cases}
 2 &\text{if } \phantom{1 < }\; \percentile(w) \leq 1\\
 \percentile(w) + 1 &\text{if } 1 < \percentile(w) < 100\\
 100 &\text{if } \phantom{1 < }\; \percentile(w) = 100
 \end{cases}
 \]
Thus, the predicted number of clusters \(k(w)\) varies between \(2\) and \(100\). 
\end{@empty}

The performance is commonly measured by two scores, the F-score and the V-measure, which capture to what extent a clustering agrees with the gold standard clustering. Since both scores are important, we rank different approaches based on the product of these scores. This automatically discounts the performance of trivial approaches: MFS, which assigns each occurrence to the same cluster, and 1cl1inst, which assigns each occurrence to its own cluster. Table~\ref{semeval:results} shows the results for OPN with different clustering algorithms and different parameters.  For comparison, the table moreover includes the best performing models of the SemEval task, as well as some other models published since.  Our best performing set-up (OPN with dbscan, $n=5000$) achieves the second best results, outperforming much more complex methods. Note that, unlike \citet{arora:linear} and \citet{mu-etal-2017-representing}, we do not use any additional data to train embeddings.

For OPN with \(k\)-means clustering, we found that \(k=30\) gives the best results among possible fixed values for \(k\).  As Table~\ref{semeval:results} shows, the performance of our method with the TPS-informed variable value \(k(w)\) is better than the performance with this fixed value.  This provides further evidence to our claim that TPS is positively correlated with the true number of meanings.  A comparison of the performance of OPN with dbscan and of OPN with \(k\)-means indicates that the size \(n\) of the neighbourhood to be considered for clustering needs to be an order of magnitude larger when we do not incorporate any information from TPS.  Our interpretation is that TPS witnesses the disturbance that an additional meaning causes in a small neighbourhood of a word, even when no word related to that meaning is present in the neighbourhood.
 
In Table~\ref{tab:evalByWord}, we single out the three best performing and the three worst performing target words with our best performing model and give the associated scores as an illustration. 

\newcommand{\ubold}[1]{\fontseries{b}\selectfont#1}
\begin{table*}[ht]
  \renewrobustcmd{\bfseries}{\fontseries{b}\selectfont}
  \renewrobustcmd{\boldmath}{}
  \newrobustcmd{\B}{\boldmath}
  \colorlet{OPNcolor}{blue!5}
  \centering
  \begin{tabular}{llccc}
    \toprule
    \textbf{Method}             & \textbf{Parameters} & {\textbf{V-Measure}}    & {\textbf{F-Score}}  & {\textbf{Product}}              \\
    \midrule
    UoY \cite{UoY} & & 0.157 & 0.498 &  \textbf{0.0782}\\
    \rowcolor{OPNcolor}
    OPN with dbscan & $n=5000$ & 0.175 & 0.420 & 0.0735 \\
    \rowcolor{OPNcolor}
    OPN with dbscan & $n=2000$ & 0.135 & 0.493 & 0.0666 \\
    \cite{mu:geometry} & $k=5$ & 0.145 & 0.441 & 0.0639 \\
    \rowcolor{OPNcolor}
    OPN with \(k\)-means & $n = 500$, $k=k(w)$ & 0.165 & 0.356 & 0.0588\\
    KSU KDD \cite{KSU:KDD} & & 0.157 & 0.369 & 0.0579 \\
    \rowcolor{OPNcolor}
    OPN with \(k\)-means & $n=500$, $k=30$ & 0.161 & 0.352 & 0.0567 \\ 
    \cite{arora:linear} & $k=5$ & 0.115 & 0.464 & 0.0533 \\
    \cite{mu:geometry} & $k=2$ & 0.073 & 0.571 & 0.0417\\
    \rowcolor{OPNcolor}
    OPN with dbscan  & $n=500$ & 0.070 & 0.571 & 0.0400 \\
    \cite{arora:linear} & $k=2$ & 0.061 & 0.586 & 0.0357\\
    \midrule
    1cl1inst &  &\textbf{0.317} & 0.090 & 0.0285 \\ 
    MFS & & 0.000 &\textbf{0.634} & 0.0000\\
    \bottomrule
  \end{tabular}%
  \caption{Performance of different methods on task~14 of SemEval-2010.  According to our ranking by product of V-measure and F-score, the algorithms UoY and KSU KDD were the strongest contenders in the initial challenge.  The algorithms MFS and 1cl1inst in the last two rows are trivial baseline algorithms}
  \label{semeval:results}
\end{table*}

\begin{table*}[ht]
\centering
\begin{tabular}{lccccccc}
\toprule
\rule[-3ex]{0pt}{7ex}\textbf{Target word} & \textbf{F-Score} & \textbf{Precision} & \textbf{Recall} & \textbf{V-Measure} & \parbox{\widthof{\textbf{geneity}}}{\hspace{0pt}\centering\textbf{Homogeneity}} & \parbox{\widthof{\textbf{Complete}}}{\hspace{0pt}\centering\textbf{Completeness}} & \textbf{Product}   \\
\midrule
presume.v       & 0.827  & 0.957     & 0.728  & 0.477     & 0.683       & 0.366        & 0.3945 \\
cultivate.v     & 0.657  & 0.648     & 0.667  & 0.518     & 0.564       & 0.479        & 0.3403 \\
accommodate.v   & 0.465  & 0.476     & 0.455  & 0.605     & 0.777       & 0.495        & 0.2813 \\
\vdots          &        &           &        &           &             &              &          \\
violate.v       & 0.153  & 0.813     & 0.085  & 0.023     & 0.292       & 0.012        & 0.0035 \\
root.v          & 0.574  & 0.405     & 0.984  & 0.000     & 0.000       & 1.000        & 0.0000 \\
sniff.v         & 0.453  & 0.295     & 0.969  & 0.000     & 0.000       & 1.000        & 0.0000  \\
\bottomrule
\end{tabular}

\caption{The performance of our best solution to SemEval on the three best performing vs.\ the three worst performing words, as evaluated according to the product of F-score and V-measure}
\label{tab:evalByWord}
\end{table*}

\section{Conclusion}
In this work, we challenge the manifold hypothesis for static word vector embeddings and experimentally show that it is more accurate and helpful to view the space of word embeddings as a pinched manifold.  We introduce a topological measure of polysemy that correlates well with the number of meanings of a word according to the gold standard of the SemEval-2010 task on \textit{Word Sense Induction \& Disambiguation}. We also produce a surprisingly simple, but topologically motivated solution to the task itself that achieves highly competitive results.

We stress that our measure of polysemy, TPS, is computed solely on the topology of the point cloud consisting of the vectors of a fixed word vector embedding.  
Of course, any solution to the described SemEval task will also predict, in particular, the number of meanings of the target words.  However, these predictions rely on access to the underlying corpus, or at least parts thereof.  
Similarly, the first step (clustering) of our solution to the SemEval task is performed directly on the word vectors, without recourse to any corpus.  This is in sharp contrast with early clustering approaches to word sense disambiguation such as \cite{schutze-1998-automatic} (which of course had to rely on far less sophisticated word vector embeddings than are now available).

A number of avenues could be pursued to further improve the results presented here.  To allow a fair comparison with other solutions to the SemEval task, we have used word vector embeddings trained on a fairly small corpus.  We have used only \emph{degree zero} persistent homology. Our method of taking the Wasserstein norm of a persistance diagram is rather crude.  
The elimination of noise from the embeddings could also  improve the results.

We conjecture that other NLP tasks that also rely, implicitly or explicitly, on the manifold hypothesis could similarly benefit from a more refined topological analysis.


\section*{Acknowledgments}
We thank Claudius Zibrowius for Figures~\ref{fig:spaces}, \ref{fig:pinched-space} and~\ref{fig:main-picture} and Peter Arndt, Michael Heck and Carel van Niekerk for helpful discussions. The results of this publication are part of the project DYMO, which has received funding from the European Research Council under the grant agreement no.~STG2018 804636. Computational resources were provided by Google Cloud.

\bibliography{anthology,main}
\bibliographystyle{acl_natbib}

\end{document}